\documentclass[11pt]{article}

\usepackage[final]{acl}

\usepackage{times}
\usepackage{latexsym}

\usepackage[T1]{fontenc}

\usepackage[utf8]{inputenc}

\usepackage{microtype}

\usepackage{inconsolata}

\usepackage{graphicx}

\usepackage{float}
\usepackage{graphicx}%
\usepackage{multirow}%
\usepackage{amsmath,amssymb,amsfonts}%
\usepackage[mathscr]{euscript}
\usepackage[title]{appendix}%
\usepackage{xcolor}%
\usepackage{textcomp}%
\usepackage{manyfoot}%
\usepackage{booktabs}%
\usepackage{algorithm}%
\usepackage{algorithmicx}%
\usepackage{algpseudocode}%
\usepackage{listings}%

\usepackage[export]{adjustbox}

\usepackage{tcolorbox}  
\usepackage{arydshln}
\newtcolorbox{AIbox}[2][]{title=#2,#1}
\usepackage{xcolor}
\definecolor{DarkGreen}{RGB}{0,100,0}
\usepackage{url}

\title{Measuring the Cost of Variety Conflation in Multilingual MT Evaluation:  Adding Mozambican Xichangana, Nyanja and Sena to FLORES+}

\author{Felermino D. M. A. Ali\textsuperscript{1}, Delfina Lázaro Mateus\textsuperscript{2}, Manuel Valente Mangue\textsuperscript{2} \\
\textsuperscript{1}Faculdade de Engenharia da Universidade Lúrio, \\Bairro Eduardo Mondlane, Pemba, Mozambique \\
\textsuperscript{2}Universidade Eduardo Mondlane - Escola de Comunicação e Artes, \\ Av. Samora Machel, Maputo, Mozambique \\
felermino.ali@unilurio.ac.mz
}

\begin{document}
\maketitle

\begin{abstract}
In this paper, we extend FLORES+ with Portuguese-source evaluation sets for three Mozambican Bantu
varieties: Xichangana, Mozambican Nyanja, and Sena. We compare Xichangana
with the existing Tsonga reference and Mozambican Nyanja with Chichewa, and
evaluate NLLB-200, Google Translate, GPT, and a variant-aware NLLB model.
Holding system output fixed reveals substantial reference sensitivity. On
\textit{devtest}, changing only the reference from Tsonga to Xichangana
reduces spBLEU by 13.10 points for NLLB-200 and 15.30 for Google. On matched
Nyanja subsets, replacing Chichewa with Mozambican Nyanja produces smaller
but consistent reductions of 3.03 and 6.10 spBLEU, respectively.
Variant-aware fine-tuning reverses this pattern on the intended targets:
relative to NLLB-200, it improves Xichangana by 7.04 spBLEU and Mozambican
Nyanja by 5.33 on \textit{devtest}, while losing performance on the sibling
references. GPT is competitive on Tsonga and Chichewa but substantially
weaker on the Mozambican varieties. For Sena, the finetuned model reaches
12.64 spBLEU and 36.21 chrF++ on \textit{devtest}. These findings motivate
variety-aware language identifiers, references, and reporting for
cross-border languages or language dialects/variants. The data is publicly available on Hugging Face at \url{https://huggingface.co/datasets/MOZNLP/FLORES_MOZ}.
\end{abstract}

\section{Introduction}

\begin{figure}[t]
\centering
\includegraphics[
  width=0.4\textwidth
]{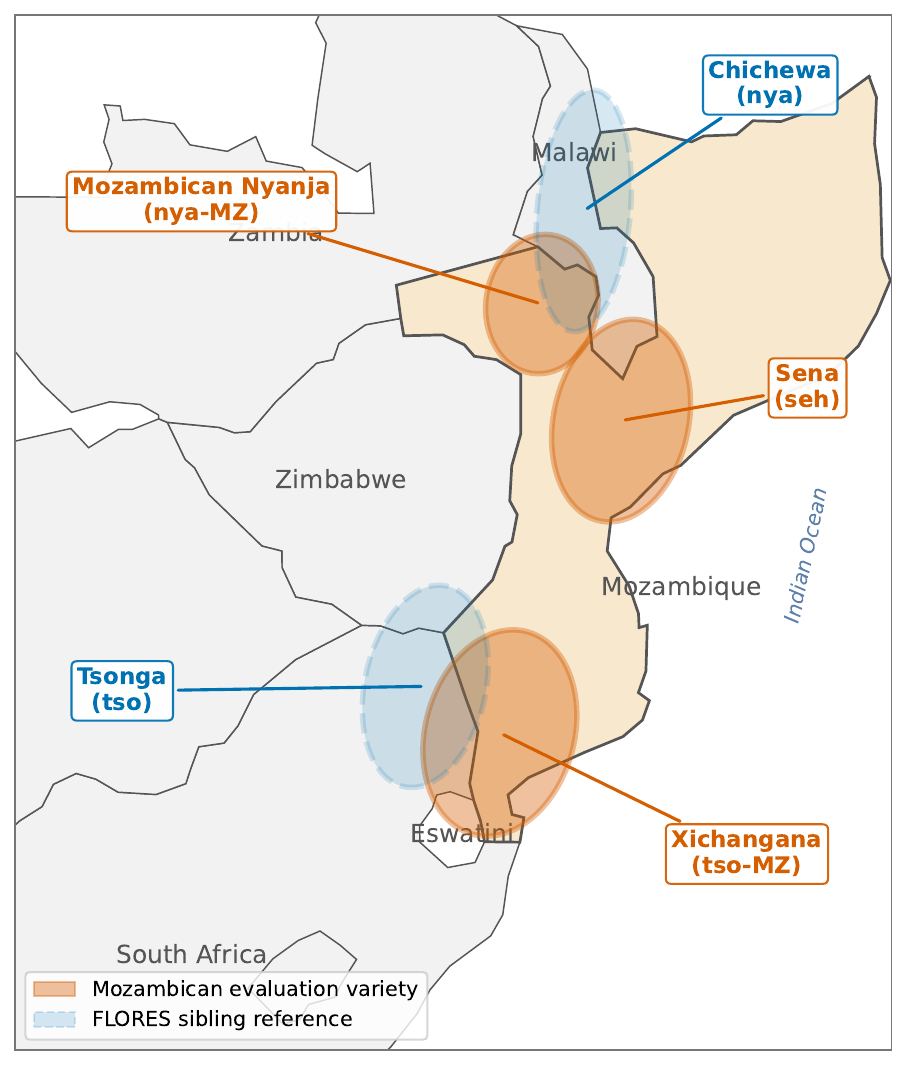}
\caption{Locations of the evaluated Mozambican varieties and their FLORES sibling references. 
}
\label{fig:target-language-map}
\end{figure}

Bantu languages are spoken across wide, contiguous regions of Africa, and colonial-era
borders rarely coincide with linguistic ones. A single language is frequently spoken in
several countries, developing varieties that differ in lexicon, orthography and, above
all, in the languages they borrow from. These differences determine whether a speaker can
actually use a translation system, and whether an evaluation score is a meaningful
estimate of that usability.

Lexical borrowing is the clearest illustration. Malawian Chichewa and the Nyanja of Tete
and Niassa in Mozambique share the code \texttt{nya}, but the former borrows
overwhelmingly from English and the latter  borrow from both Portuguese and English. A speaker in Tete reading
\textit{matilikula}, \textit{munisipiyo} or \textit{dhowutori} encounters forms that are at best
unfamiliar, where \textit{regisita}, \textit{tawoni chipi} and \textit{dokota} would be expected.
The same asymmetry holds between South African Tsonga and Mozambican Xichangana, both
tagged \texttt{tso}. Because FLORES+ contains only the South African and Malawian
varieties, systems evaluated on those sets are credited for output a Mozambican user would
not accept --- the argument recently made for Sudanese Arabic by
\citet{samil-adelani-2026-sudanese}.

We contribute (i) FLORES+ dev and devtest sets for \textbf{Xichangana} and
\textbf{Mozambican Nyanja}, translated from Portuguese by native speakers of those
varieties; (ii) the first FLORES+ set for \textbf{Sena} (\texttt{seh\_Latn}), a
two-million-speaker language of the Zambezi valley almost absent from public corpora;
(iii) a controlled comparison showing the existing \texttt{tso\_Latn} and
\texttt{nya\_Latn} sets are not valid proxies for the Mozambican varieties; and (iv) a
fine-tuning experiment showing the gap is a variety effect and is cheaply addressable.
With Emakhuwa \citep{ali-etal-2024-expanding}, these sets bring Mozambican coverage in
FLORES+ from one language to four.

\section{Related Work}

FLORES progressed from two low-resource language pairs
\citep{guzman-etal-2019-flores} to 101 languages
\citep{goyal-etal-2022-flores} and subsequently 200 languages
\citep{nllb2022}, while establishing a multi-stage
translation--review--adjudication protocol. FLORES+ is now maintained through
community governance under the Open Language Data Initiative
\citep{oldi2024}.

Recent contributions have expanded coverage, introduced language varieties,
and audited existing references. \citet{kalejaiye-etal-2025-ibom} add four
minority Nigerian languages and demonstrate their poor coverage by current
models. \citet{ali-etal-2024-expanding} contribute Emakhuwa, previously the
only Mozambican language in FLORES+, using Portuguese as the source and
documenting orthographic instability. We follow a similar protocol while
adding Sena and two Mozambican varieties.

Our variety-focused contribution is closest to the Sudanese Arabic extension
of \citet{samil-adelani-2026-sudanese}. However, because FLORES+ already
contains sibling references for Tsonga and Chichewa, we can directly measure
how evaluation changes when the reference represents the intended Mozambican
variety and whether targeted fine-tuning reduces the mismatch.

\section{The Three Languages}
\label{sec:context}

Portuguese is the official language of Mozambique but the first language of a minority;
most of the population speaks one of roughly twenty Bantu languages natively.
 \textbf{Xichangana} (\texttt{tso}) is spoken by approximately 1.8M people in Gaza, Maputo
and Inhambane, and belongs to the Tswa--Ronga group with South African Xitsonga  (see Figure~\ref{fig:target-language-map}), from
which it diverges in borrowing source, orthographic convention, and administrative
register. \textbf{Nyanja} (\texttt{nya}) is spoken by approximately 1.2M people in Tete
and Niassa, continuous with Chichewa in Malawi and Chinyanja in Zambia. \textbf{Sena}
(\texttt{seh}) is spoken by approximately 2M people along the lower Zambezi; unlike the
others it has its own code and no FLORES+ entry at all, and it is the least standardised
of the three in writing, with religious translation the dominant written genre.

\section{Approach}

We follow the FLORES+ contribution guidelines \citep{oldi2024} so that the resulting sets
are directly comparable to existing entries.

\subsection{Source data}

We translate the \textit{dev} (997 sentences) and \textit{devtest} (1{,}012 sentences)
splits of FLORES+ from Portuguese (\textit{pt}) into each of the three target varieties.
Portuguese was chosen as source rather than English for two reasons. Practically, our
translators are proficient in Portuguese and their native language but not uniformly in
English; Portuguese is the language of schooling and administration in Mozambique and thus
the language in which bilingual professional translators for these varieties actually
exist. Following \citet{ali-etal-2024-expanding}, who translated Emakhuwa from Portuguese
for the same reason, we treat translator proficiency as a stronger determinant of
reference quality than source-language canonicity. 

Because FLORES+ is multi-parallel, the Portuguese side is itself a professional
translation of the same English source used by every other entry, so our sets remain
sentence-aligned with \texttt{tso\_Latn} and \texttt{nya\_Latn}. This is what makes the
controlled comparison in Section~\ref{sec:results} possible: any score difference is
attributable to the target variety alone. Pivoting risks propagating Portuguese-side
artefacts, which we revisit in Limitations. Translators saw full document context and
worked exclusively from the Portuguese side.

\subsection{Target variety and translators}

For each language, we collaborated with two language experts, all of whom were trained linguists by the Bantu Linguistics Department at Eduardo Mondlane University. Translators had to be native speakers of the targeted variety,
resident where it is spoken, professionally competent in Portuguese, and experienced in
Portuguese-to-target translation. All
contributors were paid at or above local professional rates.

\subsection{Translation platform}

We developed a custom web-based computer-assisted translation (CAT)\footnote{CATfrica: \url{https://mozcat.vercel.app}} platform 
to support translation, quality control, and review (see Figure \ref{fig:CAT-tool}). The platform was
mobile-friendly and allowed translators to work offline, with completed work
and local changes synchronized automatically when an Internet connection
became available. This was particularly important for participants working in
settings with intermittent connectivity.
Translators were assigned individual segments and given access to bilingual
Portuguese--target-language dictionaries. For each source segment, the CAT
platform automatically identified and highlighted potentially relevant
glossary entries, helping translators select appropriate terminology and
maintain consistency.
The platform also performed automatic quality-assurance checks and displayed
warnings for potential misspellings, mismatched punctuation or numbers,
possible untranslated text, disproportionate source--target lengths, and
inconsistent capitalization. Translators could inspect each warning, revise
the translation where necessary, or confirm that the flagged difference was
intentional.
Completed segments were assigned to reviewers for independent verification.
Reviewers could correct an issue directly or return the segment to the
translator with feedback. Returned segments remained open until the translator
submitted a revision and the reviewer approved it. A segment was considered
complete only after this review cycle had been successfully concluded.

\section{Reference-Variety Analysis}
\label{sec:variant-analysis}

Before evaluating MT systems, we compare each Mozambican reference with the
corresponding FLORES sibling variety.  The analysis covers
all 2,009 aligned segments: 997 from \textit{dev} and 1,012 from
\textit{devtest}. Because the references were translated independently,
dissimilarity may reflect variety, orthographic convention, lexical choice,
paraphrase, or data quality.

\begin{table}[t]
\centering
\small
\setlength{\tabcolsep}{3.5pt}
\begin{adjustbox}{max width=\columnwidth}
\begin{tabular}{lrrrrr}
\toprule
\textbf{Reference pair} & \textbf{Segments} & \textbf{Char.\ sim.} &
\textbf{Token overlap} & \textbf{Length ratio} & \textbf{$<0.5$ length} \\
\midrule
Nyanja MZ / Chichewa & 2,008 & 54.39 & 14.33 & 1.05 & 0 \\
Xichangana / Tsonga  & 2,008 & 48.97 &  9.47 & 0.80 & 40 \\
\bottomrule
\end{tabular}
\end{adjustbox}
\caption{Reference-pair diagnostics. The Nyanja MZ/Chichewa row uses only
trusted assignee-B segments; the Xichangana/Tsonga row uses the complete
splits. Similarity and overlap are percentages; length ratio is the
sentence-level ratio of Mozambican-reference tokens to sibling-reference
tokens. The final column counts Mozambican segments with fewer than half as
many tokens as the sibling reference.}
\label{tab:reference-diagnostics}
\end{table}

\subsection{Mozambican Nyanja and Chichewa}

Mozambican Nyanja and Chichewa are related but
far from textually interchangeable. Mean character similarity is 54.39\%,
median character similarity is 54.98\%, mean token-set overlap is 14.33\%,
and vocabulary Jaccard overlap is 17.37\%. No sentence pair is an exact
match. The corpora remain broadly balanced in length with a
mean sentence-level token-length ratio of 1.05. No Mozambican segment
contains fewer than half as many tokens as its Chichewa counterpart.
The observed divergence is therefore not explained by systematic
truncation.

The strongest recurring distinction is the treatment of \textit{c} and
\textit{ch}. A bare \textit{c} not followed by \textit{h} occurs 1,610 times
in the Mozambican reference but 367 times in Chichewa; \textit{ch} occurs
281 and 1,849 times, respectively. The most frequent aligned orthographic
correspondences include \textit{ca}/\textit{cha} and
\textit{nchito}/\textit{ntchito}, each observed 12 times, followed by
\textit{awili}/\textit{awiri} and \textit{nzinda}/\textit{mzinda}.
Additional examples include \textit{cifukwa}/\textit{chifukwa},
\textit{caka}/\textit{chaka}, and
\textit{mwacitsanzo}/\textit{mwachitsanzo}.

The difference is not reducible to spelling. Recurrent substitutions such as
\textit{zimene}/\textit{zomwe}, \textit{amene}/\textit{omwe}, and
\textit{monga}/\textit{ngati} indicate differences in lexical and
grammatical selection. Names and borrowed forms also vary, as illustrated by
\textit{afilika}/\textit{africa}.

\subsection{Xichangana and Tsonga}

Xichangana and Tsonga show a stronger separation. Mean character similarity
is 48.94\%, median character similarity is 49.47\%, mean token-set overlap is
9.47\%, and vocabulary Jaccard overlap is 9.87\%. Again, no sentence pair is
an exact match. The Xichangana reference contains 40,893 tokens, compared
with 52,443 in Tsonga, yielding a mean token-length ratio of 0.80.

The most prominent orthographic distinction is nearly complementary:
\textit{sv} occurs 3,681 times in Xichangana and only 5 times in Tsonga,
whereas \textit{sw} occurs 42 and 4,191 times, respectively. Examples include
\textit{sva}/\textit{swa}, \textit{lesvaku}/\textit{leswaku},
\textit{lesvi}/\textit{leswi}, and
\textit{nasvona}/\textit{naswona}. Other recurring correspondences include
\textit{la}/\textit{ra}, \textit{doropa}/\textit{doroba},
\textit{ndzhaku}/\textit{endzhaku},
\textit{nkama}/\textit{nkarhi}, and
\textit{mpimu}/\textit{mpimo}. Function words and agreement forms also vary,
including alternations among \textit{na}/\textit{ni},
\textit{ka}/\textit{eka}/\textit{ku}, and several noun-class agreement
markers.

\section{Experimental Setup}
\label{sec:experimental-setup}

We evaluate three systems for Portuguese-to-target translation. The
\textbf{NLLB-200} baseline is
\texttt{facebook/nllb-200-distilled-600M}. \textbf{Google} denotes the Google
Cloud Translation general NMT model\footnote{Queried on September, 2026}, with
Portuguese specified as the source and \texttt{ny} or \texttt{ts} as the
target. Google provides no separate identifiers for the Mozambican varieties,
so each Google output is scored against both the sibling and Mozambican
reference. No Google Sena output was collected.

\textbf{GPT}\footnote{\texttt{gpt-5.6-sol}, queried on September, 2026} denotes openAI API endpoints. A zero-shot translation prompt (see Appendix \ref{sec:gpt-prompts})
specified Portuguese as the source and separately described standard
Chichewa, Mozambican Nyanja, standard Xitsonga/Tsonga, Xichangana, and Sena as
the five targets. Each response was constrained to structured JSON and
validated for complete segment and target coverage. GPT therefore generates a
distinct output for every target variety, including Sena, rather than sharing
one hypothesis across each sibling--Mozambican pair.

\textbf{NLLB+finetuned} starts from the same distilled 600M NLLB checkpoint.
We retain the standard \texttt{nya\_Latn}, \texttt{tso\_Latn}, and
\texttt{seh\_Latn} tokens for Chichewa, Tsonga, and Sena, and introduce
\texttt{nya\_MZ\_Latn} and \texttt{tso\_MZ\_Latn} for Mozambican Nyanja and
Xichangana. The two new token embeddings are initialized when the tokenizer
vocabulary is extended and learned during fine-tuning. We also prepend an
explicit target marker such as \texttt{<2nya-MZ>} or \texttt{<2tso-MZ>} to each
source sentence. The fine-tuned model can therefore generate different
hypotheses for the sibling and Mozambican targets.

\subsection{Fine-tuning data}

\begin{table*}[t] \centering \small \setlength{\tabcolsep}{3.5pt} \begin{adjustbox}{max width=\textwidth} \begin{tabular}{lp{10cm}r} \toprule \textbf{Target variety} & \textbf{Corpus/domain coverage} & \textbf{Train (pt$\leftrightarrow$XX)} \\ \midrule Chichewa (\texttt{nya}) & Jehovah Witness & 56,902 \\ Nyanja (\texttt{nya-MZ}) & Portuguese news: culture, sport, economy & 8,164 \\ Tsonga (\texttt{tso}) & Jehovah Witness & 54,678 \\ Xichangana (\texttt{tso-MZ}) & Jehovah Witness; UDHR; Legal; Portuguese news: culture, sport, economy & 43,662 \\ Sena (\texttt{seh}) & Jehovah Witness; Portuguese news: culture, sport, economy & 64,532 \\ \bottomrule \end{tabular} \end{adjustbox} \caption{Data coverage. Counts are directed sentence-pair rows for Portuguese--variety translation directions.} \label{tab:finetuning-data} \end{table*}

Table~\ref{tab:finetuning-data} reports the amount of direct Portuguese data
and the total number of eligible rows involving each target variety, together
with the corpus or domain coverage represented in the training data.

\subsection{Optimization}

Training uses four NVIDIA RTX A6000 GPUs with DistributedDataParallel.

The model is trained for 15,000 optimizer
steps with maximum sequence length
200. We use Adafactor with learning rate $1\times10^{-4}$, weight decay
$1\times10^{-3}$, clip threshold 1.0, and a constant schedule after
250 warmup steps. Random seeds are initialized from 222 with a rank-specific
offset. Model, tokenizer, optimizer, and scheduler states are saved every
500 steps. 

\subsection{Evaluation}

The aligned evaluation data contain 997 \textit{dev} and 1,012
\textit{devtest} sentences per direction. We translate the complete
Portuguese source into \texttt{nya}, \texttt{nya-MZ}, \texttt{tso},
\texttt{tso-MZ}, and, where supported, \texttt{seh}.

For the off-the-shelf NLLB baseline, the sibling and Mozambican labels map to
the same standard NLLB tokens, so its hypotheses are identical within each
pair and only the reference changes. Google similarly produces one
\texttt{ny} output and one \texttt{ts} output per Portuguese sentence.
NLLB+finetuned instead uses the distinct regional tokens described above.

We recompute all scores from the saved hypotheses using SacreBLEU 2.6.0.
spBLEU uses the \texttt{flores200} tokenizer\footnote{\textit{nrefs:1|case:mixed|eff:no|tok:flores200|smooth:exp}}; chrF++ uses
character order 6 and word order 2\footnote{\textit{nrefs:1|case:mixed|eff:yes|nc:6|nw:2|space:no}}. Results are reported
separately for \textit{dev} and \textit{devtest}.

\section{Results and Discussion}
\label{sec:results}

\begin{table*}[t]
\centering
\scriptsize
\setlength{\tabcolsep}{2.5pt}
\begin{adjustbox}{max width=\textwidth}
\begin{tabular}{lcccccccccc}
\toprule
\textbf{System} & \multicolumn{5}{c}{\textbf{Dev}} & \multicolumn{5}{c}{\textbf{Devtest}} \\
\cmidrule(lr){2-6}\cmidrule(lr){7-11}
 & \textbf{pt$\rightarrow$tso} & \textbf{pt$\rightarrow$tso-MZ} & \textbf{pt$\rightarrow$nya} & \textbf{pt$\rightarrow$nya-MZ} & \textbf{pt$\rightarrow$seh} & \textbf{pt$\rightarrow$tso} & \textbf{pt$\rightarrow$tso-MZ} & \textbf{pt$\rightarrow$nya} & \textbf{pt$\rightarrow$nya-MZ} & \textbf{pt$\rightarrow$seh} \\
\midrule
NLLB-200 & 20.67 / 46.12 & 5.15 / 30.74 & 15.21 / 42.60 & 10.41 / 39.08 & -- & 18.91 / 44.82 & 5.81 / 32.98 & 14.70 / 42.12 & 12.56 / 39.70 & -- \\
Google &\textbf{ 23.94 / 48.84 }& 5.01 / 31.57 & \textbf{18.08 / 45.81} & 10.11 / 39.54 & -- & \textbf{22.42 / 47.89} & 7.11 / 36.03 & \textbf{18.16 / 45.65} & 12.37 / 39.96 & -- \\
GPT & 22.17 / 48.30 & 7.05 / 33.77 & 15.07 / 44.84 & 7.12 / 33.21 & 9.26 / 34.46 & 21.39 / 47.68 & 7.46 / 34.55 & 15.14 / 44.75 & 8.71 / 33.52 & 9.92 / 35.30 \\
NLLB+finetuned & 17.42 / 42.30 & \textbf{10.34 / 36.45} & 10.76 / 36.81 & \textbf{15.87 / 44.13} & 10.56 / 34.07 & 17.50 / 42.05 &\textbf{ 12.85 / 38.95 }& 11.06 / 37.35 & \textbf{17.89 / 44.98} & \textbf{12.64 / 36.21} \\
\bottomrule
\end{tabular}
\end{adjustbox}
\caption{Portuguese-to-XX results on \texttt{dev} and \texttt{devtest}. Each cell reports spBLEU / chrF++.}
\label{tab:pt-flores-flat-dev-devtest}
\end{table*}

Table~\ref{tab:pt-flores-flat-dev-devtest} reports spBLEU and chrF++ on the
Portuguese-source evaluation data. The Tsonga, Xichangana, Chichewa, and Sena
columns use the complete splits: 997 \textit{dev} and 1,012
\textit{devtest} sentences.

On the complete sibling-language sets, Google is the strongest system. On
\textit{devtest}, it obtains 22.42 spBLEU / 47.89 chrF++ for Tsonga and
18.16 / 45.65 for Chichewa. GPT is close on Tsonga at 21.39 / 47.68 and
reaches 15.14 / 44.75 on Chichewa, while NLLB-200 obtains
18.91 / 44.82 and 14.70 / 42.12, respectively. The ranking changes on the
Mozambican targets. NLLB+finetuned is strongest for Xichangana at
12.85 / 38.95 and for Mozambican Nyanja at 17.89 / 44.98. It also
provides the baseline Sena result on \textit{devtest}, reaching
12.64 / 36.21.

\subsection{Sibling references substantially overestimate Xichangana quality}
\label{sec:varietygap}

Xichangana exhibits a large and consistent mismatch with the existing FLORES
Tsonga reference. On \textit{devtest}, the same NLLB-200 hypothesis scores
18.91 / 44.82 against Tsonga but only 5.81 / 32.98 against Xichangana,
reductions of 13.10 spBLEU and 11.84 chrF++. Google shows the same pattern,
falling from 22.42 / 47.89 to 7.11 / 36.03, with reductions of 15.30 spBLEU and
11.87 chrF++.

The mismatch is larger on \textit{dev}. NLLB-200 falls from
20.67 / 46.12 on Tsonga to 5.15 / 30.74 on Xichangana, while Google falls
from 23.94 / 48.84 to 5.01 / 31.57. The corresponding differences are
15.53 and 18.93 spBLEU and 15.38 and 17.27 chrF++, respectively. Because
these comparisons change only the reference, the gaps cannot be attributed
to decoding, prompting, or model selection. They demonstrate that
performance measured on \texttt{tso\_Latn} is not a reliable proxy for
performance on Mozambican Xichangana.

This finding is consistent with the reference analysis in
Section~\ref{sec:variant-analysis}. Xichangana and Tsonga exhibit low token
overlap, systematic orthographic differences such as \textit{sv} versus
\textit{sw}, and substantial lexical and grammatical divergence. A system
can consequently receive a strong Tsonga score while failing to produce
forms appropriate for Xichangana. 

\subsection{Mozambican Nyanja evaluation}

Google obtains 10.11 / 39.54 on \textit{dev} and 12.37 / 39.96 on
\textit{devtest}, close to NLLB-200 at 10.41 / 39.08 and
12.56 / 39.70. GPT is weaker, reaching 7.12 / 33.21 and
8.71 / 33.52.

In \textit{devtest}, NLLB-200 decreases from
15.59 / 42.80 in Chichewa to 12.56 / 39.70 in Mozambican Nyanja, resulting in reductions
of 3.03 spBLEU and 3.10 chrF++. Google decreases from 18.47 / 45.88 to
12.37 / 39.96, reductions of 6.10 spBLEU and 5.92 chrF++. On
\textit{dev}, NLLB-200 decreases from 14.38 / 42.61 to 10.41 / 39.08,
while Google decreases from 17.19 / 45.62 to 10.11 / 39.54.

The matched data therefore reveal a systematic Chichewa--Mozambican Nyanja
gap, but one smaller than the Xichangana--Tsonga gap.

\subsection{Variant-aware fine-tuning}
\label{sec:finetune}

Variant-aware fine-tuning produces substantial gains on both Mozambican
targets. On the complete Xichangana \textit{devtest} set, NLLB+finetuned
improves over NLLB-200 from 5.81 / 32.98 to 12.85 / 38.95, gains of
7.04 spBLEU and 5.97 chrF++. It also exceeds Google at 7.11 / 36.03 and GPT
at 7.46 / 34.55. On \textit{dev}, it reaches 10.34 / 36.45, compared with
5.15 / 30.74 for NLLB-200, 5.01 / 31.57 for Google, and
7.05 / 33.77 for GPT.

On the Mozambican Nyanja subsets, NLLB+finetuned improves over
NLLB-200 from 12.56 / 39.70 to 17.89 / 44.98 on \textit{devtest}, gains of
5.33 spBLEU and 5.28 chrF++. It exceeds Google by 5.52 spBLEU and
5.02 chrF++, and GPT by 9.18 spBLEU and 11.46 chrF++. On \textit{dev}, it
obtains 15.87 / 44.13, compared with 10.41 / 39.08 for NLLB-200,
10.11 / 39.54 for Google, and 7.12 / 33.21 for GPT. The results therefore
show a clear fine-tuning advantage on both splits.

The Mozambican gains coincide with lower scores on the complete sibling
sets. On \textit{devtest}, fine-tuning changes Tsonga from
18.91 / 44.82 to 17.50 / 42.05 and Chichewa from 14.70 / 42.12 to
11.06 / 37.35. The same trade-off appears on \textit{dev}, where Tsonga
falls from 20.67 / 46.12 to 17.42 / 42.30 and Chichewa from
15.21 / 42.60 to 10.76 / 36.81. This pattern is consistent with the model
learning variety-specific lexical and orthographic choices rather than
receiving a uniform improvement across related targets. Because the adapted
system generates separate hypotheses for each target identifier, the result
demonstrates controllable specialization rather than a reference
substitution effect.

\subsection{Sena, and broader implications}

GPT performs competitively on the established sibling languages but is much
weaker on the Mozambican targets. On \textit{devtest}, its Tsonga result
(21.39 / 47.68) approaches Google, and its Chichewa result
(15.14 / 44.75) exceeds NLLB-200 in chrF++. However, GPT reaches only
7.46 / 34.55 on Xichangana and 8.71 / 33.52 on the Mozambican
Nyanja subset. This contrast suggests that specifying a variety in a
zero-shot prompt does not provide sufficient control for these
underrepresented targets.

For Sena, NLLB+finetuned obtains 10.56 / 34.07 on \textit{dev} and
12.64 / 36.21 on \textit{devtest}, compared with 9.26 / 34.46 and
9.92 / 35.30 for GPT. Fine-tuning yields higher Sena spBLEU on both splits
and higher chrF++ on \textit{devtest}, although GPT is marginally higher in
chrF++ on \textit{dev}. Since no NLLB-200 or Google Sena output is
available, this comparison covers the two systems with complete Sena
predictions but does not establish improvement over the original NLLB
baseline.

Overall, the results support following conclusions: (1) sibling references
can substantially overestimate translation quality for Mozambican varieties,
especially Xichangana. (2) supervised variety-aware fine-tuning is substantially more effective than generic
decoding or zero-shot target naming on the Mozambican targets.

We therefore report sibling and Mozambican references separately. Automatic
metrics establish reference sensitivity and the value of target-specific
adaptation, but they cannot determine whether an output is acceptable to the
intended speakers.

\section{Conclusion}

We contribute Portuguese-source FLORES+ evaluation sets for Xichangana,
Mozambican Nyanja, and Sena. The results show that an existing reference in a
closely related national variety is not necessarily an adequate proxy for
the community being evaluated. Under identical hypotheses, replacing the
Tsonga reference with Xichangana reduces \textit{devtest} spBLEU by
13.10 points for NLLB-200 and 15.30 for Google. The corresponding
Chichewa--Mozambican Nyanja reductions on matched evaluation subsets are
smaller but remain systematic, at 3.03 and 6.10 spBLEU. These differences
are consistent with the orthographic, lexical, and grammatical patterns
observed in the reference analysis and demonstrate that aggregate language
codes can conceal meaningful variety-level differences.

Variant-aware NLLB fine-tuning substantially improves performance on the
intended Mozambican targets. On \textit{devtest}, it gains 7.04 spBLEU and
5.97 chrF++ over NLLB-200 for Xichangana, and 5.33 spBLEU and 5.28 chrF++
for Mozambican Nyanja. These gains coincide with lower scores on Tsonga and
Chichewa, indicating target-specific specialization rather than a uniform
model improvement. GPT remains competitive on the established sibling
languages but performs markedly worse on the Mozambican varieties, while
the adapted model establishes a reproducible Sena result of
12.64 spBLEU / 36.21 chrF++ on \textit{devtest}. We therefore recommend
variety-specific references, target identifiers, and disaggregated reporting
for cross-border Bantu languages so that benchmark scores correspond to the
communities that MT systems are intended to serve.

\section{Limitations}

Our references were produced from Portuguese rather than the original English source.
Although the Portuguese side is itself a professional translation and the sets remain
sentence-aligned with all other entries, pivoting risks propagating Portuguese-specific
phrasing and existing errors into our targets; an English-sourced control subset is left
to future work. 

\section{Ethical Considerations}

All translators were professionals compensated at or above prevailing local rates, gave
informed consent for release.

\section*{Acknowledgements}
This work was carried out with support from \url{lacunafund.org} and \url{google.org}. \textbf{Disclaimer}: The views expressed herein do not necessarily represent those of Lacuna Fund, its Steering Committee, or CENIA.


\bibliography{anthology}

@inproceedings{ali-etal-2024-expanding,
    title = "Expanding {FLORES}+ Benchmark for More Low-Resource Settings: {P}ortuguese-Emakhuwa Machine Translation Evaluation",
    author = "Ali, Felermino Dario Mario  and
      Lopes Cardoso, Henrique  and
      Sousa-Silva, Rui",
    editor = "Haddow, Barry  and
      Kocmi, Tom  and
      Koehn, Philipp  and
      Monz, Christof",
    booktitle = "Proceedings of the Ninth Conference on Machine Translation",
    month = nov,
    year = "2024",
    address = "Miami, Florida, USA",
    publisher = "Association for Computational Linguistics",
    url = "https://aclanthology.org/2024.wmt-1.45/",
    doi = "10.18653/v1/2024.wmt-1.45",
    pages = "579--592"
}

@inproceedings{guzman-etal-2019-flores,
    title = "The {FLORES} Evaluation Datasets for Low-Resource Machine Translation: {N}epali{--}{E}nglish and {S}inhala{--}{E}nglish",
    author = "Guzm{\'a}n, Francisco  and
      Chen, Peng-Jen  and
      Ott, Myle  and
      Pino, Juan  and
      Lample, Guillaume  and
      Koehn, Philipp  and
      Chaudhary, Vishrav  and
      Ranzato, Marc{'}Aurelio",
    editor = "Inui, Kentaro  and
      Jiang, Jing  and
      Ng, Vincent  and
      Wan, Xiaojun",
    booktitle = "Proceedings of the 2019 Conference on Empirical Methods in Natural Language Processing and the 9th International Joint Conference on Natural Language Processing (EMNLP-IJCNLP)",
    month = nov,
    year = "2019",
    address = "Hong Kong, China",
    publisher = "Association for Computational Linguistics",
    url = "https://aclanthology.org/D19-1632",
    doi = "10.18653/v1/D19-1632",
    pages = "6098--6111",
}

@article{goyal-etal-2022-flores,
  title = {The {FLORES}-101 Evaluation Benchmark for Low-Resource and Multilingual Machine Translation},
  author = {Goyal, Naman and Gao, Cynthia and Chaudhary, Vishrav and Chen, Peng-Jen and Wenzek, Guillaume and Ju, Da and Krishnan, Sanjana and Ranzato, Marc'Aurelio and Guzm\'an, Francisco and Fan, Angela},
  journal = {Transactions of the Association for Computational Linguistics},
  volume = {10},
  pages = {522--538},
  year = {2022}
}

@article{nllb2022,
  title = {No Language Left Behind: Scaling Human-Centered Machine Translation},
  author = {{NLLB Team}},
  journal = {arXiv preprint arXiv:2207.04672},
  year = {2022}
}

@misc{oldi2024,
  title = {Open Language Data Initiative: Contribution Guidelines},
  author = {{Open Language Data Initiative}},
  year = {2024},
  howpublished = {\url{https://oldi.org}}
}

@inproceedings{kalejaiye-etal-2025-ibom,
  title = {Ibom {NLP}: A Step Toward Inclusive Natural Language Processing for {N}igeria's Minority Languages},
  author = {Kalejaiye, Oluwadara and Beyene, Luel Hagos and Adelani, David Ifeoluwa and Edet, Mmekut-Mfon Gabriel and Akpan, Aniefon Daniel and Urua, Eno-Abasi and Andy, Anietie},
  booktitle = {Proceedings of the 2025 Conference of the Asia-Pacific Chapter of the Association for Computational Linguistics (IJCNLP-AACL)},
  year = {2025},
  url = {https://aclanthology.org/2025.ijcnlp-long.22/}
}

@inproceedings{samil-adelani-2026-sudanese,
  title = {Sudanese-Flores: Extending {FLORES}+ to {S}udanese {A}rabic Dialect},
  author = {Samil, Hadia Mohmmedosman Ahmed and Adelani, David Ifeoluwa},
  booktitle = {Proceedings of the 7th Workshop on African Natural Language Processing (AfricaNLP)},
  address = {Rabat, Morocco},
  pages = {243--247},
  publisher = {Association for Computational Linguistics},
  year = {2026},
  url = {https://aclanthology.org/2026.africanlp-main.25/}
}

\appendix

\clearpage

\clearpage

\section{Reference-Variety Examples}
\label{app:variant-examples}

Table~\ref{tab:variant-examples} presents two aligned examples from the
evaluation data. For each example, it shows the English reference, the
Portuguese source used for translation, the Mozambican Nyanja and Chichewa
references, and the Xichangana and Tsonga references. Selected forms are
highlighted to make recurring orthographic and lexical contrasts easier to
identify.

\begin{table}[t]
\centering
\small
\setlength{\tabcolsep}{5pt}
\begin{adjustbox}{max width=\textwidth}
\begin{tabular}{lp{0.84\textwidth}}
\toprule
\textbf{Language} & \textbf{Reference sentence} \\
\midrule
& \\
\multicolumn{2}{l}{\textbf{Example 1:} ``Photons are even smaller than the
stuff that makes up atoms!''} \\
\addlinespace[2pt]
\textbf{Portuguese source} &
Os fótons são ainda menores do que as partículas que formam os átomos! \\
\cmidrule(lr){1-2}
Mozambican Nyanja &
Fotoni tili \textbf{tocepa} kwambiri kuposera mbali
\textbf{zimene} zimapanga matomo! \\
Chichewa &
Ma photon ndi \textbf{ocheperapo} kuposa zinthu
\textbf{zomwe} zimapanga ma atom. \\
\cmidrule(lr){1-2}
Xichangana &
\textbf{Mafotoni} mahali matsongo kutlula
\textbf{sviphemu lesvi sviwumbaka} ma-atomo! \\
Tsonga &
\textbf{Tifothoni} i titsongo eka
\textbf{swilo leswi} endlaka tiathomu! \\

\midrule
\midrule

& \\
\multicolumn{2}{l}{\textbf{Example 2:} ``It's worth half an hour to stroll
about the intriguing village.''} \\
\addlinespace[2pt]
\textbf{Portuguese source} &
Vale a pena caminhar meia hora pela intrigante vila. \\
\cmidrule(lr){1-2}
Mozambican Nyanja &
Ndi kosangalatsa kwambiri kuyenda hafu \textbf{awala} muvila
\textbf{yocititsa} kasoyo. \\
Chichewa &
Ndikofunikila theka la \textbf{ola} kuuyendela mudzi
\textbf{wopatsa chidwiwo}. \\
\cmidrule(lr){1-2}
Xichangana &
\textbf{Svayampsa} kufambafamba hafu ya wara hi mugaga
\textbf{wohlamalisa}. \\
Tsonga &
\textbf{Swi fanele} awara na hafu ku famba famba eka tiko xikaya
\textbf{leri ro tsakisa}. \\
\bottomrule
\end{tabular}
\end{adjustbox}
\caption{Aligned examples showing the English reference, Portuguese source,
Mozambican varieties, and their FLORES sibling references. Boldface
highlights selected orthographic and lexical contrasts;}
\label{tab:variant-examples}
\end{table}

\clearpage
\section{GPT translation prompts}
\label{sec:gpt-prompts}

The following system and user prompts were used for GPT translation. In the
user prompt, \texttt{<INPUTS\_JSON>} represents the JSON serialization of a
batch of Portuguese source segments, each containing its split, segment
identifier, and source text. Long lines below are wrapped for typesetting;
the wrapped portions are joined with spaces in the actual prompt.

\paragraph{System prompt.}
\begin{quote}
\begin{minipage}{0.96\linewidth}
\small
\begin{verbatim}
You are a professional African-language translator. Translate faithfully
from Portuguese. Preserve meaning, names, numbers, punctuation, and
sentence boundaries. Produce natural target-language text, not
explanations. Keep regional varieties distinct. Return valid JSON only.
\end{verbatim}
\end{minipage}
\end{quote}

\paragraph{User prompt.}
\begin{quote}
\begin{minipage}{0.96\linewidth}
\small
\begin{verbatim}
Translate every input into all five targets.

Target definitions:
- "nya": standard Chichewa as used in Malawi; use standard Malawian
  Chichewa vocabulary and orthography
- "nya-MZ": Mozambican Nyanja; use vocabulary and orthographic
  conventions appropriate for Nyanja speakers in Mozambique
- "tso": standard Xitsonga/Tsonga as used in South Africa; use standard
  South African Xitsonga vocabulary and orthography
- "tso-MZ": Xichangana as used in Mozambique; use Mozambican vocabulary
  and orthography rather than South African Xitsonga
- "seh": Sena as used in Mozambique; use natural Sena vocabulary and
  orthography

Return exactly this JSON shape:
{"translations":[{"split":"dev","segment_id":1,"nya":"...",
"nya-MZ":"...","tso":"...","tso-MZ":"...","seh":"..."}]}

Return one object per input, in the same order. Do not omit, merge, or add
segments. Do not include Markdown.

Inputs:
<INPUTS_JSON>
\end{verbatim}
\end{minipage}
\end{quote}

Each input object in \texttt{<INPUTS\_JSON>} has the following form:
\begin{quote}
\begin{minipage}{0.96\linewidth}
\small
\begin{verbatim}
{"split":"dev","segment_id":1,"portuguese":"..."}
\end{verbatim}
\end{minipage}
\end{quote}

\clearpage
\onecolumn
\section{The CATfrica Translation Tool}
\label{app:catfrica}

CATfrica is a web-based Computer-Assisted Translation (CAT) tool built to support
translation work for low-resource languages. It covers the whole cycle of a translation job:
project preparation, task distribution, translation, quality control, revision,
delivery and payment.

\paragraph{Roles and workflow.}
Five roles are supported: administrator, project manager, translator, revisor and
annotator. A project passes through the stages \emph{draft}, \emph{in progress},
\emph{under review}, \emph{approved} and \emph{paid}, and each sentence carries its own
status, so different parts of a document can be assigned to different translators and
progress can be monitored precisely. Large documents can be split automatically into
sub-projects translated in parallel.

\paragraph{Project preparation.}
Managers upload one or more text files and choose sentence- or paragraph-level
segmentation. The original layout is preserved, so the translated document can be
rebuilt with the same paragraphs, spacing and punctuation. Translators may split or
merge segments when the automatic division is inadequate.

\paragraph{Translation editor.}
Source and target are shown side by side. For every sentence the tool automatically
offers previous translations of identical or similar sentences from the project's
translation memories, terminology suggestions from its glossaries, and machine
translation. It further provides search and replace, concordance search, per-sentence
comments, version history and automatic saving. If the connection is lost, work
continues offline and is uploaded automatically when connectivity returns.

\paragraph{Quality assistance.}
A spell checker flags words absent from the target-language word list and proposes
corrections. Automatic checks warn about unusual length differences, unbalanced
brackets or quotes, missing or altered numbers, inconsistent final punctuation, double
spaces, wrong apostrophes and stray line breaks. Both components were adapted to the
orthographic conventions of African languages; word-internal apostrophes, for
instance, are not treated as errors.

\paragraph{Memories, glossaries and annotation.}
Translation memories and glossaries can be created in the tool or imported from
spreadsheets and attached to any project. Current version support biligual dictionary for Portuges--- Nyanja, Xichangana, Emakhuwa and Sena, and English --- Swahili, Xhosa. Matching is tolerant: suggestions appear for
merely similar sentences, and terms are recognised in inflected forms. Terms may be
labelled as ordinary terms, loanwords, code-switching or bilingual expressions. An
alignment function additionally lets annotators link a source expression to its
counterpart in the translation, so that annotated bilingual lexicons and
code-switching resources are produced alongside the translation itself.

\begin{figure}[h]
    \centering
    \includegraphics[width=1\linewidth]{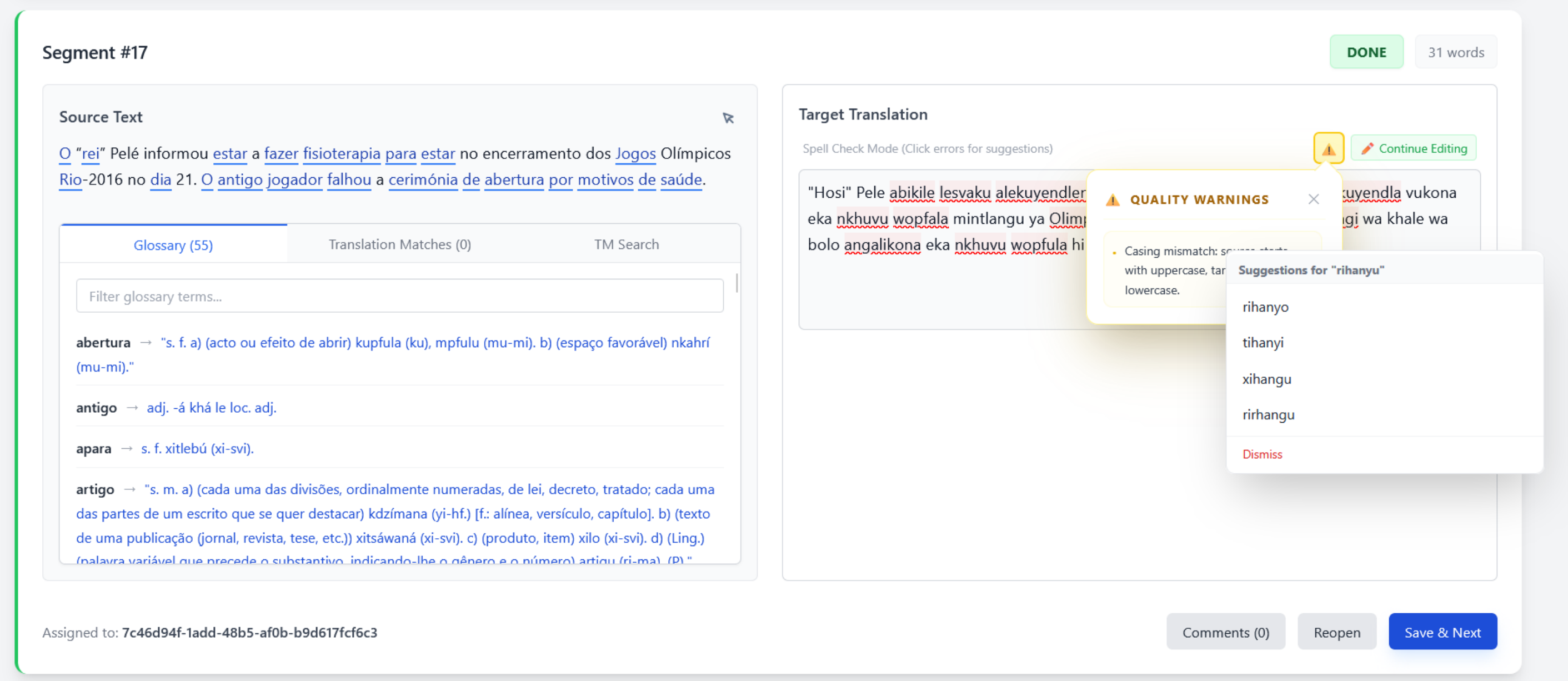}
    \caption{Screenshot of the custom web-based computer-assisted translation (CAT) platform
to support translation. }
    \label{fig:CAT-tool}
\end{figure}

\paragraph{Machine translation.}
Each project can be linked to an engine adapted to its language pair. Suggestions are
reused when the same sentence recurs, and models can alternatively be downloaded and
run on the translator's own computer, which matters where connectivity is limited.

\paragraph{Revision and delivery.}
Finished translations are reviewed sentence by sentence by a revisor, who cannot be the
original translator; rejected sentences return to the translator for correction.
Approved projects are exported as formatted text, as an archive with one file per
uploaded document, as structured data including the alignments, or in a standard
translation-memory format for reuse as a resource or as training data.

\paragraph{Payment and administration.}
Managers set a per-word rate; earnings are computed from completed words and settled
through invoices. Users purchase word credits via PayPal or M-Pesa, with all
transactions recorded in a transparent statement. An administration area manages
accounts and invitations, supported languages, translation engines and backups.

\end{document}